\documentclass[letterpaper, 10 pt, conference]{ieeeconf}  
\pdfoutput=1

\IEEEoverridecommandlockouts                              

\usepackage{times} 

\usepackage[ruled,linesnumbered]{algorithm2e}
\usepackage{algpseudocode}

\usepackage{amsmath} 
\usepackage{amssymb}  
\usepackage{color}
\usepackage{graphicx}
\usepackage{subfigure}
\usepackage[english]{babel}
\usepackage{amsthm}
\usepackage{breqn}
\usepackage[super]{nth}
\usepackage{url}
\usepackage{siunitx}
\usepackage{makecell}
\usepackage{microtype}
\usepackage{booktabs}
\usepackage{hyperref}
\usepackage{stfloats}
\usepackage{threeparttable} 
\usepackage{multirow}
\theoremstyle{definition}

\usepackage{verbatim}

\usepackage[utf8]{inputenc}
\usepackage{graphicx}
\usepackage{subfigure}
\usepackage{float}
\usepackage{bm}
\usepackage{amssymb}
\usepackage{threeparttable}
\usepackage{multirow}
\usepackage{eqnarray}
\usepackage{color}
\usepackage{graphicx}
\usepackage{tablefootnote}

\makeatletter
\newcommand{\removelatexerror}{\let\@latex@error\@gobble}
\makeatother

\SetKwInOut{KwParameter}{Parameters}

\title{\LARGE \bf
Efficient Model Learning and Adaptive Tracking Control of Magnetic Micro-Robots for Non-Contact Manipulation
}
\author{
Yongyi Jia, Shu Miao, Junjian Zhou, Niandong Jiao, Lianqing Liu, and 
Xiang Li
\thanks{Y. Jia, S. Miao, and X. Li are with the Department of Automation, Tsinghua University, China. J. Zhou, N. Jiao and L. Liu are with the State Key Laboratory of
Robotics, Shenyang Institute of Automation, Chinese Academy of Sciences,
Shenyang 110016, China. 
This work was supported in part by Tsinghua University Initiative Scientific Research Program, in part by Beijing National Research Center for Information Science and Technology, in part by the National Natural Science Foundation of China under Grant U21A20517 and 52075290, and in part by the Science and Technology Innovation 2030-Key Project under Grant 2021ZD0201404. Corresponding author: Xiang Li (xiangli@tsinghua.edu.cn)}
}

\begin{document}

\maketitle
\thispagestyle{empty} 
\pagestyle{empty}  

\newtheorem{definition}{Definition}

\begin{abstract}

Magnetic microrobots can be navigated by an external magnetic field to autonomously move within living organisms with complex and unstructured environments. Potential applications include drug delivery, diagnostics, and therapeutic interventions. Existing techniques commonly impart magnetic properties to the target object, or drive the robot to contact and then manipulate the object, both probably inducing physical damage. This paper considers a non-contact formulation, where the robot spins to generate a repulsive field to push the object without physical contact. Under such a formulation, the main challenge is that the motion model between the input of the magnetic field and the output velocity of the target object is commonly unknown and difficult to analyze. To deal with it, this paper proposes a data-driven-based solution. A neural network is constructed to efficiently estimate the motion model. Then, an approximate model-based optimal control scheme is developed to push the object to track a time-varying trajectory, maintaining the non-contact with distance constraints. Furthermore, a straightforward planner is introduced to assess the adaptability of non-contact manipulation in a cluttered unstructured environment. Experimental results are presented to show the tracking and navigation performance of the proposed scheme.
\end{abstract}

\section{Introduction}
Robotics micromanipulation \cite{rufo2022acoustofluidics,wang2019intracellular,dai2019robotic} is a prominent and active research area, which holds paramount significance across an array of applications within disciplines, including biomedicine, electronics, and materials science \cite{ren20193d,ren2021plasmonic}. Many strategies have been applied to micromanipulation automation, such as {mechanical \cite{zhang2016robotic}, fluidic \cite{miao2023microfluidics}, electrical \cite{fan2011electric}, optical \cite{li2013dynamic} and magnetic \cite{kummer2010octomag}.}
Among them, magnetic microrobots have the attractive feature of moving within living organisms. To improve its autonomous capability, diverse methodologies have been devised to attain meticulous command over the positioning, orientation, and motion of magnetic particles \cite{xu2022independent,yang2022optimal,du2023image}. However, existing results are commonly under the formulation of direct manipulation \cite{wang2018magnetic}, imparting magnetic properties to the target object, which would disrupt its inherent properties, or contact manipulation, driving the robot to contact and then manipulate the object \cite{li2019biped}.
The introduction of physical contact would probably result in deformation, contamination, and difficulty in separating. 
\begin{figure}[th]
\vspace{2pt}
    \centering
    \includegraphics[width=8.5cm]{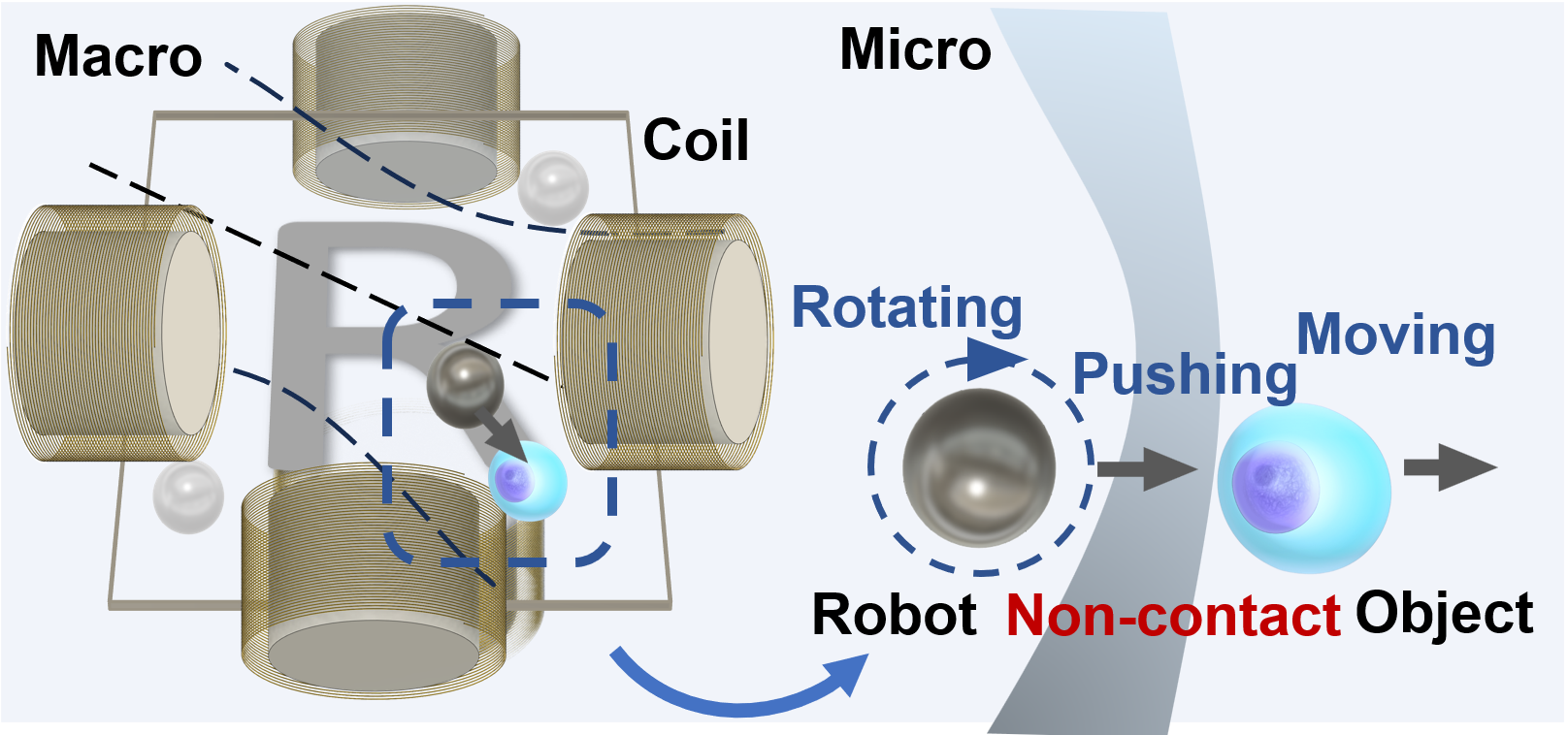}
    \vspace{-10pt}
    \caption{Illustration diagram about magnetic robot performing non-contact manipulation on the target object. The rotating magnetic field generated by the electromagnetic coil propels the robot to roll, thereby inducing fluid motion to drive the target object. 
    }
    \label{head_view}
    \vspace{-20pt}
\end{figure}
To deal with the aforementioned drawbacks, this paper will be dedicated to proposing an indirect control approach that does not involve contact. Basically, the robot rotates under a magnetic field to generate repulsive and eddy fields to push the object, as shown in Fig. \ref{head_view}. Under such a formulation,
the analysis of the motion model from magnetic field input to target velocity output is a complex fluid-structure interaction (FSI) problem that involves the bidirectional  influence between the robot, the target object, and the surrounding fluid. Furthermore, the uncertainty of dynamic parameters adds challenge to the acquisition of the motion model.

This paper introduces a data-driven solution to the aforementioned problem. While much data-driven progress has been reported to estimate the unknown model at the macro scale, it is not trivial to reproduce it at the micro scale since i) few sensory techniques, ii) stochastic Brownian motion, and especially, iii) the inefficiency or even absence of simulation methods making data collection exceedingly challenging. 

In this paper, a neural network is constructed to quickly estimate the proposed succinct model expression with a real-world collected small data set. Then, a model-based control scheme is developed to push the target object, tracking the desired time-varying trajectory. The control input is obtained by solving a convex optimization problem with distance constraints, guaranteeing real-time performance and non-contact ability. Finally, a curvature optimization-based planner is introduced to validate the navigation ability of non-contact manipulation. Experimental results are carried out to show the tracking and navigation performance of the proposed method. 

The indirect non-contact manipulation manner is similar to the non-prehensible manipulation \cite{ruggiero2018nonprehensile}, especially planar pushing \cite{moura2022non,10341476} at the macro scale, such that many newly-emerging techniques at the macro scale can also be applied to the microrobot to bridge the gap between each other.
\section{Related Works}
\vspace{-3pt}
Magnetic field is most promising for the cell transport function due to its high penetration of 
body, precise controllability and good biocompatibility \cite{202110625}.
Several works have been proposed to manipulate micro-objects in a non-contact manner under a magnetic and flow field. One approach involves propelling the robot translationally using a gradient magnetic field, thereby generating laminar flow along the robot's boundaries. 
Floyd et al. \cite{floyd2009two} introduced a method that employs micro-robots to harness the fluid velocity field generated by the fluid boundary layer, thereby inducing the non-contact motion of the target object.
Then it was expanded by Pawashe et al. in \cite{pawashe2011two} as two manipulation strategies: 1) front pushing, involving direct contact between the microrobot and the micro-object; and 2) side pushing, which enables non-contact displacement through the utilization of fluid flow fields generated by a translating microrobot. The effective range of the translational mode has limited effectiveness, making long-distance transportation of objects challenging to achieve. 


Another approach involves leveraging the eddy currents generated by the robot's rotating or rolling in a uniform magnetic field. Pieters et al. \cite{pieters2015rodbot} presented the modeling and control of a rolling microrobot, the RodBot. The RodBot is capable of executing non-contact pushing operations and lifting objects utilizing a vortex above its body. Petit et al. \cite{petit2012selective} introduced a novel approach for achieving controlled manipulation of individual micro- and nanoscale objects through the utilization of mobile micro vortices, which enabled trapping and manipulation within fluidic environments without significant constraints on material properties. Ye et al. \cite{ye2014dynamic} introduced a technique that demonstrated the ability to dynamically trap and precisely transport individual swimming microorganisms in a two-dimensional manner, all without requiring the use of additional labeling on the biological samples or specialized surface patterns. Lin et al. \cite{lin2018magnetically} presented a magnetically actuated peanut-shaped hematite colloid motor capable of both rolling and wobbling motion in fluids, which enabled single-cell non-contact transportation. This eddy non-contact mode is more flexible while achieving closed-loop control using eddy can be challenging. Therefore, Khalil et al. \cite{khalil2020controlled} proposed a non-contact manipulation strategy using a rotating two-tailed soft microrobot to induce a frequency-dependent flow-field for orbiting nonmagnetic microbeads, with a closed-loop controller to guide the microbeads towards target positions.

Our work is inspired by recent research \cite{dai2022magnetically}, which introduced a conceptual diagram along with experimental demonstrations depicting a non-contact object manipulation scenario by the cell robot. Specifically, during rotation, hydrodynamic pressure at the robot's edge effectively pushes the object forward. Compared to the previous two non-contact methods, this approach utilizing local pressure proves to be more stable and easier to control. Therefore, we employ this non-contact mode based on pushing, enabling precise closed-loop control and automatic navigation.

In summary, non-contact indirect manipulation offers advantages such as the absence of physical damage, reduced contamination risk, and ease of separation, which have garnered the attention of researchers. However, there are relatively few works regarding the closed-loop control and navigation of non-contact manipulation, which are either controlled by humans or applicable to setpoint control only, limiting its applications. 
\vspace{-2pt}
\section{Indirect Non-contact Manipulation}
\vspace{-2pt}
In this section, we explore models of non-contact indirect magnetic drive systems. Section III. A presents the principle and steady-state analysis of the one-dimensional scenario, illustrating how magnetic robots manipulate objects without contact. Section III. B extends this to two-dimensional motion, introducing dynamic equations for the following control.
\begin{figure*}[htbp]
\centering
\includegraphics[width=15.5cm]{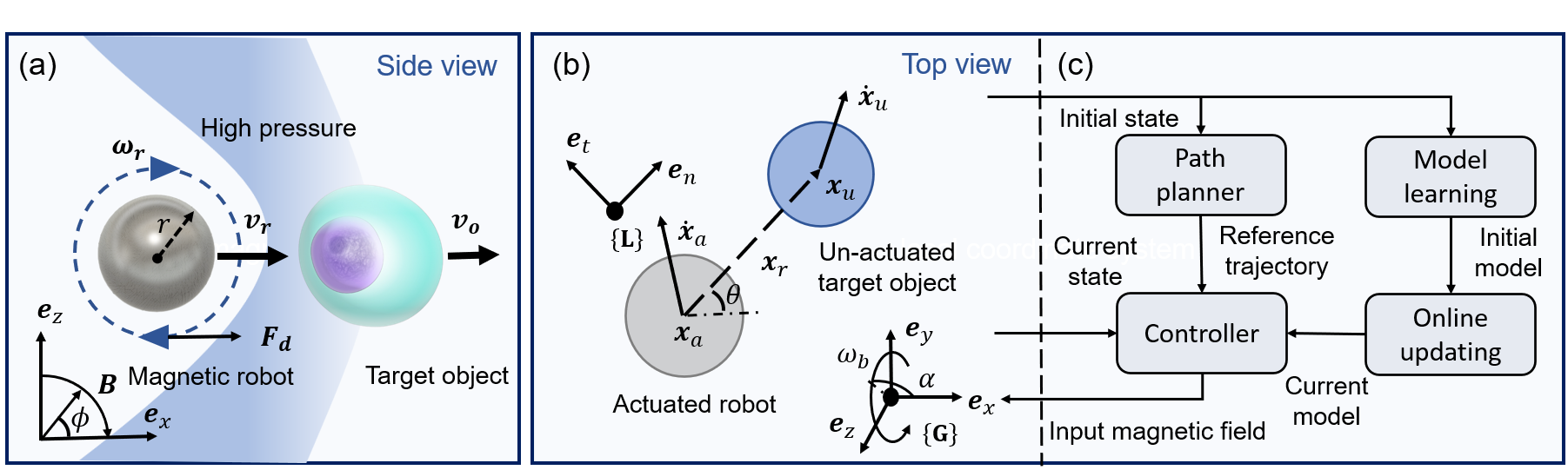}
\vspace{-0.2cm}
\caption{Overview of the proposed non-contact manipulation scheme. \textbf{(a)} One-dimensional schematic diagram of non-contact pushing.
\textbf{(b)} Two-dimensional motion model employing a local coordinate system, where the normal and the tangential are coupled. 
\textbf{(c)} Control structure of non-contact manipulation. The controller receives the reference trajectory from the planner and the estimation model based on the neural network, then gives the control input through optimization and online updating.
}
\label{2}
\vspace{-20pt}
\end{figure*}
\subsection{Linear Non-Contact Motion}
As illustrated in Fig. \ref{2}(a), a spherical robot with rolling motion is driven by a rotating magnetic field in the x-z plane. 
A distinctive rotating uniform magnetic field is characterized by an angular velocity $\omega_b$ and an amplitude $B$, expressed as 
\vspace{-2pt}
\begin{equation}
    \bm{B}(\omega_b,t) = B \cos (\omega_bt + \theta) \bm{e}_x + B \sin (\omega_bt + \theta) \bm{e}_z,
     \label{eq: rotation magnetic}
\end{equation}
\hspace{-2pt}
where $\bm{e}_x$ and $\bm{e}_z$ denote the unit vectors along the x-axis and z-axis, respectively, and $\theta$ represents the initial phase angle. Due to the non-uniform magnetic sputtering on the surface, the robot microsphere has a magnetic moment $\bm{m}$ and then is subjected to a magnetic torque, which can be represented as
\vspace{-2pt}
\begin{equation}
    {\tau}_m = \bm{m} \times \bm{B} = mB \sin(\Phi),
     \label{eq: magnetic torque}
\end{equation}
where $\Phi$ is the angle between the magnetic moment and the magnetic field. The equilibrium between magnetic torque and viscous torque infers that the robot rotates at the same angular velocity as the magnetic field, with a linear velocity proportional to the angular velocity, that is $\omega_r = \omega_b $, and $v_r = a\omega_b$, until the step-out frequency is reached \cite{sinn2011magnetically}. 

When focusing on indirect manipulation, the rotation of the robot drives surrounding fluid flow with a surface velocity as $\omega_r r$, where $r$ is the radius of the robot, significantly outpacing the robot's linear speed $v_r$, which results in high-pressure ahead the robot, facilitating conditions for indirect non-contact manipulation of objects \cite{dai2022magnetically}. Considering both the robot and target object as a single entity, the rotation and linear motions are obtained as
\vspace{-2pt}
\begin{eqnarray}
    \tau_m - C_m^r \omega_r - F_d r = 0 ,
     \label{eq: whole_rotation}\\
    F_d - C_d^r v_r - C_d^o v_o = 0 ,
     \label{eq: whole_linear}
\end{eqnarray}
where $v_o$ is the linear velocity of the target object, $F_d$ is the drag force between the robot and the underlying surface \cite{xu2022multimodal}, $C_m$ is the viscous moment coefficient, and $C_d$ is the viscous drag coefficient. As the distance between them diminishes, pressure increases until the target object's speed matches the robot's linear speed, that is, $v_r = v_o$, resulting in a constant phase angle difference $\Phi$ and a relative state of rest. The ratio of sliding and rotating $\frac{v_r}{\omega_r}$ which depends on both object property and environmental damping can be obtained by measuring the linear velocity. 
\vspace{-2.5pt}
\subsection{Planar Non-Contact Motion}
\vspace{-2.5pt}
In this paper, we focus on addressing the task of the magnetic-driven robot:  1) pushing an object along a designated trajectory and 2) navigating it to a target location in a cluttered environment within a two-dimensional plane. Both the robot and the manipulated object are assumed to be homogeneous spheres suitable for most microscopic manipulation targets. 

As shown in Fig. \ref{2}(b), the magnetic field rotates around an axis in the xy plane. The angle between its projection onto the xy plane and the x-axis is denoted as $\alpha$.  When the robot's motion direction differs from the line connecting it to the object's center, a tangential fluid force affects the object, causing its curved motion. Due to the intricate dynamics of fluid-structure interaction \cite{bertrand2009influence}, calculating these challenges remains tough, even using a multiphysics simulator such as COMSOL Multiphysics \cite{dickinson2014comsol}. Hence, we introduce a control affine model to depict the planar motion, which can be estimated through a data-driven method. 

Below, the positions of the actuated robot and un-actuated target object are denoted as $\bm{x}_a \in \mathbb{R}^2$ and $\bm{x}_u \in \mathbb{R}^2$. Since we are only concerned with the control of the target object while maintaining a non-contact state between the two objects, the system state is set to $x = \left[ \bm{x}_u,\bm{x}_r\right] ^\intercal \in \mathbb{R}^4$, where $\bm{x}_r = \bm{x}_u -\bm{x}_a$ is the relative position. The control input $\bm{u}$ is defined as $\left[ \omega_b \cos \alpha, \omega_b \sin \alpha \right] ^\intercal \in \mathbb{R}^2$, representing the projected angular velocity on the x and y-axis. 
Several properties can be noticed intuitively: 1) quasi-static: under low Reynolds coefficients, viscous resistance plays the prevailing factor and the system follows a first-order model; 2) Movement is solely induced by a magnetic field, rendering the drift vector field $f(\cdot)$ of the system as none; 3) Both the liquid and magnetic field are uniformly distributed, then the control vector field $g(\cdot)$ is solely determined by relative positioning.
Therefore, the system model can now be given as
\vspace{-3pt}
\begin{equation}
    \bm{\dot x} = \bm{g}(\bm{x}_r)\bm{u} ,
     \label{eq: whole_linear}
\end{equation}
where $\bm{g}: \mathbb{R}^2 \rightarrow \mathbb{R}^{4,2}$ is the unknown nonlinear control function about $\bm x_r$. 
\vspace{-0.1cm}

\section{Modeling Learning and Tracking Control}
This section proposes a new adaptive tracking controller to achieve the non-contact pushing task. First, a neural network is constructed to efficiently estimate the motion model. The controller is formulated as an optimization problem, such that constraints can also be added to avoid contact. A planner with curvature optimization is proposed to verify the navigation capability. The whole structure of the scheme is illustrated in Fig. \ref{2}(c). 
\vspace{-2pt}
\subsection{Model Learning}
An offline data-driven strategy is adapted for constructing the dynamic model to overcome the analytical challenges caused by fluid-structure interaction. To enhance model simplicity and reduce the data sampling requirements, we introduce a local coordinate system, serving the line connecting the robot's center and the target object's center as the normal direction reference, as shown in Fig. \ref{2}(b). The control equation exclusively depends on the relative distance. Additionally, the motion along the tangential direction and the normal direction becomes decoupled. To further improve model generalization, we scale-normalize the relative distances by the mean radius of the robot and the object. 

Then, (\ref{eq: whole_linear}) can be expressed in concise forms
\vspace{-2pt}
\begin{eqnarray}
    &\bm{\dot x}_u = \bm{g}_u\left(\bm{x}_r\right)\bm{u} = \bm{R}^\intercal\bm{N}_u\left(s_r\right)\bm{R}\bm{u} ,
     \label{eq: gu_equation}\\
    &\bm{\dot x}_r = \bm{g}_r\left(\bm{x}_r\right)\bm{u} = \bm{R}^\intercal\bm{N}_r\left({s}_r\right)\bm{R}\bm{u} ,
     \label{eq: gr_equation}
\end{eqnarray}
where $\bm{R} \in \mathbb{SO}\left(2\right)$ is the global-to-local rotation matrix for the angle $\arctan \left (\bm{x}_{rx}, \bm{x}_{ry} \right)$, $ s_r \in \mathbb{R}$ is shorthand for normalized relative Euclidean distance $\|\bm{x}_r\|_2 / (r_a + r_u)$ , and $\bm{N}_u,\bm{N}_r : \mathbb{R} \rightarrow \mathbb{R}^{2,2}$ return diagonal matrices due to direction decoupling. A unified form is used to represent the following unactuated and relative formulas for simplicity, distinguishing by subscript $k \in \{u,r\}$.

We apply radial basis function network (RBFN) to estimate the unknown nonlinear functions $\bm{N}_u$ and $\bm{N}_r$. RBFN possesses the capability for rapid learning and can be updated online \cite{yu2011advantages}, attaining significant advancements in the modeling and control of deformable objects \cite{9888782}. The actual function is expressed as 
\begin{equation}
    \bm{N}_k\left(s_r\right) = \text{diag}\left(\bm{W}_k \bm{\phi}\left(s_r\right)\right), \quad k \in \{u, r\}
     \label{eq: N_equation}
\end{equation}
where $\bm{\phi}\left(s_r\right) = \left[\phi_1\left(s_r\right), \phi_2\left(s_r\right), \cdots,\phi_p\left(s_r\right)\right]^\intercal \in  \mathbb{R}^p$ is the activation function vector shared between two networks. Each network is equipped with its own set of weights, $\bm{W}_u$ and $\bm{W}_r \in \mathbb{R}^{2,p}$, and $\text{diag}(\cdot)$ represents the operator creating diagonal matrices from vectors. Given micro-environmental motion with Brownian noise, we employ a multi-quadric radial function as the activation function. 
\begin{equation}
    \bm{\phi}_i\left(s_r\right) = \sqrt{1 + \|\sigma_i(s_r-\mu_i)\|_2^2} . \quad i = 1, 2, \cdots, p 
     \label{eq: activation function}
\end{equation}
The parameters $\mu_i$ and $\sigma_i$ are the center and width of the $i$th neuron, which are trained offline and fixed online. This activate function could maintain smoothness in input parameters while offering robustness in noisy data scenarios. 

By substituting (\ref{eq: N_equation}) to (\ref{eq: gu_equation}) and (\ref{eq: gr_equation}), the actual motion model is rewritten as 
\vspace{-2pt}
\begin{equation}
    \bm{\dot x}_k = \bm{R}^\intercal\text{diag}\left(\bm{W}_k \bm{\phi}\left(s_r\right)\right)\bm{R}\bm{u}. 
     \label{eq: actual_equation}
\end{equation}
With the estimated weight matrix $\bm{\hat W}_k$ trained from the data set, the estimated function is represented as 
\begin{equation}
    \bm{\hat g}_k(\bm{x}_r) = \bm{R}^\intercal\text{diag}\left(\bm{\hat W}_k \bm{\phi}\left(s_r\right)\right)\bm{R}. 
     \label{eq: estimated_equation}
\end{equation} 
Thus, the prediction error of the model is calculated as
\begin{equation}
    \bm{e}_k = \bm{\dot x}_k - \bm{\hat g}_k(\bm{x}_r)\bm{u} = \bm{R}^\intercal\text{diag}\left(\Delta\bm{W}_k \bm{\phi}\left(s_r\right)\right)\bm{R}\bm{u},
     \label{eq: prediction_error}
\end{equation} 
where $\Delta\bm{W}_k = \bm{W}_k - \hat {\bm{W}}_k$ is the approximation error of the RBFN weights. We manually manipulate the magnetic field's direction and angular velocity in a real environment to collect offline data. Each data tuple encompasses relative position, object velocity, relative velocity, and magnetic field input, as $\{\bm{x}_r, \bm{v}_u, \bm{v}_r, \bm{u}\}$. The data set is ensured to span a broad range of relative distances from contacting to keeping away. Mean squared error (MSE) loss of $e_k$ and Adam optimizer \cite{kingma2014adam} are used for offline training. The simplicity of the RBFN's structure enables excellent fitting performance with a minimal data requirement, thus facilitating manual data collection.  

\subsection{Adaptive Optimal Control}
After estimating the nonlinear system model, a straightforward approach involves employing feedback linearization for control. However, it is crucial to ensure that the robot maintains a safe distance from the target object to facilitate non-contact. To address this challenge, we introduce an adaptive optimal controller. A convex optimization problem is formulated to compute the control inputs while incorporating distance constraints to prevent any physical contact. Simultaneously, the controller dynamically updates the neural network weights in real time to compensate for modeling errors, thus enabling adaptation to alterations in both objects and environmental conditions.

The trajectory tracking is transformed into a series of fixed-point control problems involving a sequence of target points. The desired target position $\bm{x}_u^{\text{{des}}}$ is chosen as the next point on the trajectory, and then the target task error is defined as
\begin{equation}
\Delta \bm{x}_u = \bm{x}_u - \bm{x}_u^{\text{des}}.
\label{eq: delta}
\end{equation}
To push the target object forward along the trajectory, the desired relative position is aligned opposite to the initial target task error as
\begin{equation}
\bm{x}_r^{\text{des}} = - x_r^{\text{avg}} \frac{\bm{x}_u - \bm{x}_u^{\text{des}}}{\|{\bm{x}_u - \bm{x}_u^{\text{des}}}\|_2},
\label{eq: desire}
\end{equation}
where $x_r^{\text{avg}}$ is the average value of the relative distance in the data set corresponding to a specific range of target object speeds. The relative task error is obtained as $\Delta \bm{x}_r = \bm{x}_r - \bm{x}_r^{\text{des}}$. Then, ideal velocity vectors are defined in the opposite direction of the task errors as
\begin{eqnarray}
&{\dot{\bm{x}}}_u^{\text{ide}} = -\alpha \Delta {\bm{x}_u}, 
\label{eq: ideal_u}\\
&{\dot{\bm{x}}}_r^{\text{ide}} = -\alpha \Delta {\bm{x}_r},
\label{eq: ideal_r}
\end{eqnarray}
where $\alpha$ is a positive gain factor.

The control input is obtained by solving the following convex optimization problem
\begin{align}
    &\underset{\bm{u}}{\text{min}} \quad \frac{1}{2}\| \dot{\bm{x}}^{\text{ide}}_u - \hat{\bm{g}}_u(\bm{x}_r)\bm{u}\|^2_2 + \frac{\lambda}{2} \| \dot{\bm{x}}^{\text{ide}}_r - \hat{\bm{g}}_r(\bm{x}_r)\bm{u}\|^2_2 \notag \\
    & \text{s.t.} \quad \|\bm{u}\|_2  \leq {u}_{\text{max}} \notag \\
     &\quad \quad {\bm{x}_r}^\intercal\hat{\bm{g}}_r(\bm{x}_r)\bm{u} \leq 0 \quad \text{if} \quad s_r \leq  s_r^{\text{min}} \notag\\ 
     &\quad \quad {\bm{x}_r}^\intercal\hat{\bm{g}}_r(\bm{x}_r)\bm{u} \geq 0 \quad \text{if} \quad s_r \geq  s_r^{\text{max}} 
    \label{eq: optimal}
\end{align}
where the cost function is designed to bring the target object closer to the desired position as possible, and $\lambda$ is a tunable parameter whose increase aids in tracking trajectories with higher curvature. As mentioned above, the contact between the robot and the target object will cause solid contamination, and it is difficult to separate due to the adhesion force. Therefore, we introduce safety distance constraints to maintain a non-contact state. As indicated in the second inequality, when the distance is less than or equal to the minimum safe distance $s_r^{\text{min}}$, the projection of the relative velocity onto the relative distance becomes negative, resulting in an expansion of the relative distance. This principle also applies when the distance exceeds the maximum operating distance $s_r^{\text{max}}$. 

The online updating law of the estimated weight matrix is designed as 
\begin{align}
\dot{\hat{\bm{W}}}_{k,i}^\intercal = \lambda_k \sum \limits_{j=1}^{2} &R_{i,j}(\tau_1\Delta x_{k,j} +\tau_2 e_{k,j}) \bm{R}_i \bm{u} \bm{\phi}(s_r), \notag \\ 
&i = 1,2, \quad \lambda_k = \left \{\begin{array}{ll}
1, & k \quad = \quad u \\ 
\lambda, & k \quad = \quad r
\end{array}\right.
\label{eq: online update}
\end{align}
where $\Delta x_{k,j}$ are the $j$th element of the target and relative task error, $e_{k,j}$ are the $j$th element of the target and relative prediction error, and $\tau_1$, $\tau_2$ are positive scalars. The control rate (\ref{eq: optimal}) combined with the online update rate (\ref{eq: online update}) theoretically guarantees the Lyapunov stability of the closed-loop system \cite{slotine1991applied}.

\subsection{Navigation in Clutter Environments}
We develop a straightforward path planning method for guiding a target object to its destination, demonstrating the application of our proposed automated non-contact manipulation in real clutter micro-environments. Addressing the curvature limitations imposed by non-holonomic constraints on the target object's trajectory, we propose the Curvature-Optimized RRT (CO-RRT) algorithm, an enhancement to the Rapidly-exploring Random Tree (RRT) framework. 

First, the turning angle is considered in the distance calculation when checking the nearest node of the randomly sampled one. The turning angle is defined as
\begin{equation}
\theta  = \arccos \left(\frac{\bm{x}_{{c}}\cdot \bm{x}_{{p}}}{\|\bm{x}_{{c}}\|_2 \|\bm{x}_{{p}}\|_2}\right),
\label{eq: turn angle}
\end{equation}
where $\bm{x}_{{c}}$ is the position of current node and $\bm{x}_{{p}}$ is the position of parent node. The distance measure is designed by the summation of the orientation difference and the Euclidean distance as 
\begin{equation}
d(\bm{p}_{{r}}, \bm{p}_{t}) = \left |\arccos\left(\frac{\text{tr}\left(\bm{R}_{{r}}^\intercal\bm{R}_{{t}}\right)}{2}\right) \right| + \|\bm{x}_{{r}} - \bm{x}_{{t}}\|_2,
\label{eq: distance}
\end{equation}
where the subscripts ${r}, {t}$ respectively represent randomly sampled nodes and nodes on the expanding tree. $\bm{R}_{r}$ and $\bm{R}_{t}$ are the rotation matrices of turning angles $\theta_r$ and $\theta_t$.

After obtaining a set of reference trajectory points $\bm{x}_1^{\text{des}}, \bm{x}_2^{\text{des}},  \cdots , \bm{x}_n^{\text{des}}$, we employ Bézier curves \cite{prautzsch2002bezier} to smooth the trajectory. The continuous reference trajectory along the Bézier curve is depicted as
\begin{equation}
\bm{x}^{\text{des}}(t) = \sum_{i=0}^{n} b_i^q(t)\bm{x}_i,
\label{eq: Bezier}
\end{equation}
where $b_i^q(t)$ are the Bézier basis functions defined as
\begin{equation}
b_i^q(t) = \frac{q!}{i!(q-i)!} (1-t)^{q-i}t^i.
\label{eq: Bezier basis}
\end{equation}
The degree $q$ of the Bezier curve determines its smoothness and complexity. This approach diminishes the maximum curvature of the planned trajectory, consequently leading to a substantial reduction in tracking errors.

\section{Experiments}
\begin{figure}[ht]
    \centering
    \vspace{0pt}
     \includegraphics[width=6.8cm]{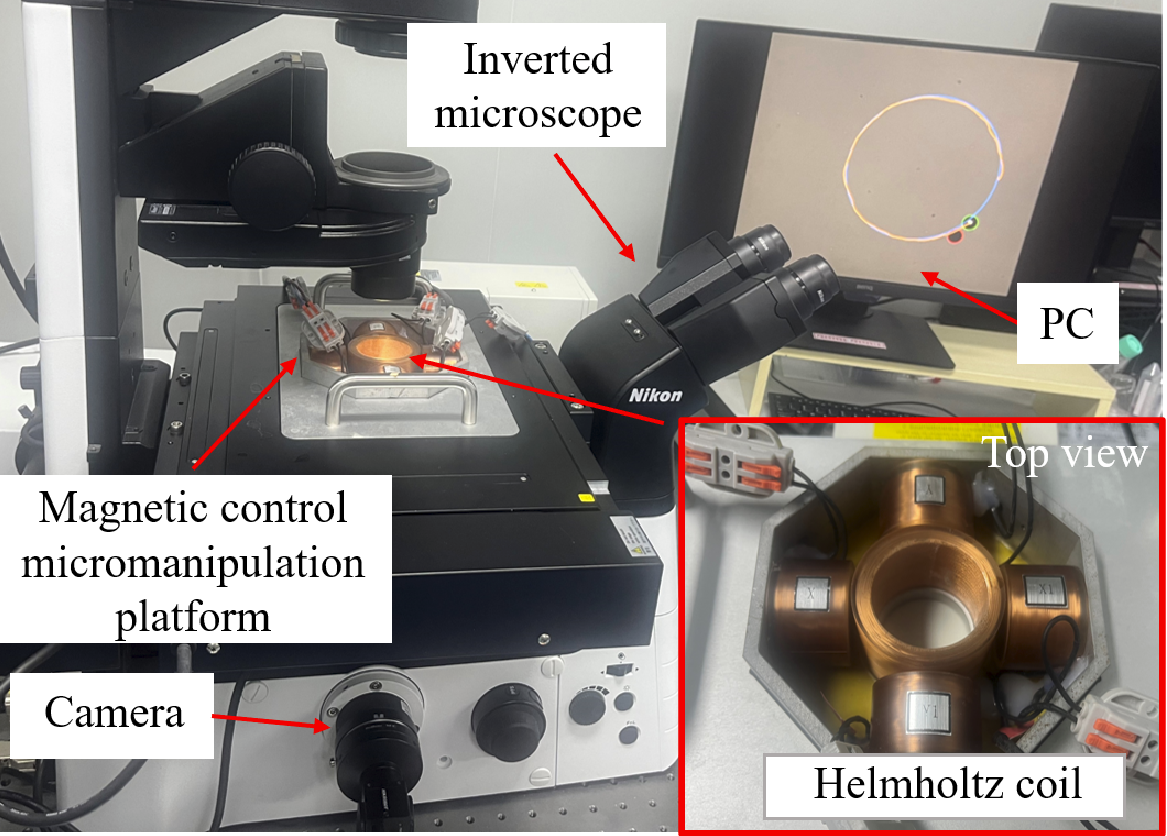}
     \vspace{-5pt}
    \caption{System setup. The three-axis Helmholz coil generates a rotating magnetic field, while a camera is connected to an inverted microscope to capture images. The magnetic field inputs are calculated on the PC side.}
    \label{setup}
    \vspace{-5pt}
\end{figure}

As shown in Fig. \ref{setup}, the system consists of a self-designed three-axis Helmholtz electromagnetic coil mounted on the motorized stage of the inverted microscope (Nikon, ECLIPSE Ti2-E), in which the specimen is embedded via a 3D printed support structure. A CMOS camera (HIKROBOT, MV-CH050-10UM) is connected to the inverted microscope to capture images.
On the PC side, OpenCV \cite{bradski2000opencv} is employed for the recognition and localization of the magnetic robot and target object. Subsequently, control inputs are computed and transmitted to a servo amplifier (Maxon, ESCON 50/5) to drive the electromagnetic coil. The amplitude of the magnetic field is maintained at 10 mT, and the angular velocity and direction of rotation are calculated by the controller. The magnetic robot is fabricated by spin-coating a soft magnetic nickel layer onto silica microspheres. Using 10 um microspheres, one hour of DC magnetron sputtering results in a nickel coating thickness of approximately 480 nm. The target objects consist of untreated 10 um microspheres.
\subsection{Validation of Model Learning Method}
\begin{figure}[ht]
    \centering
    \vspace{-10pt}
    \includegraphics[width=6cm]{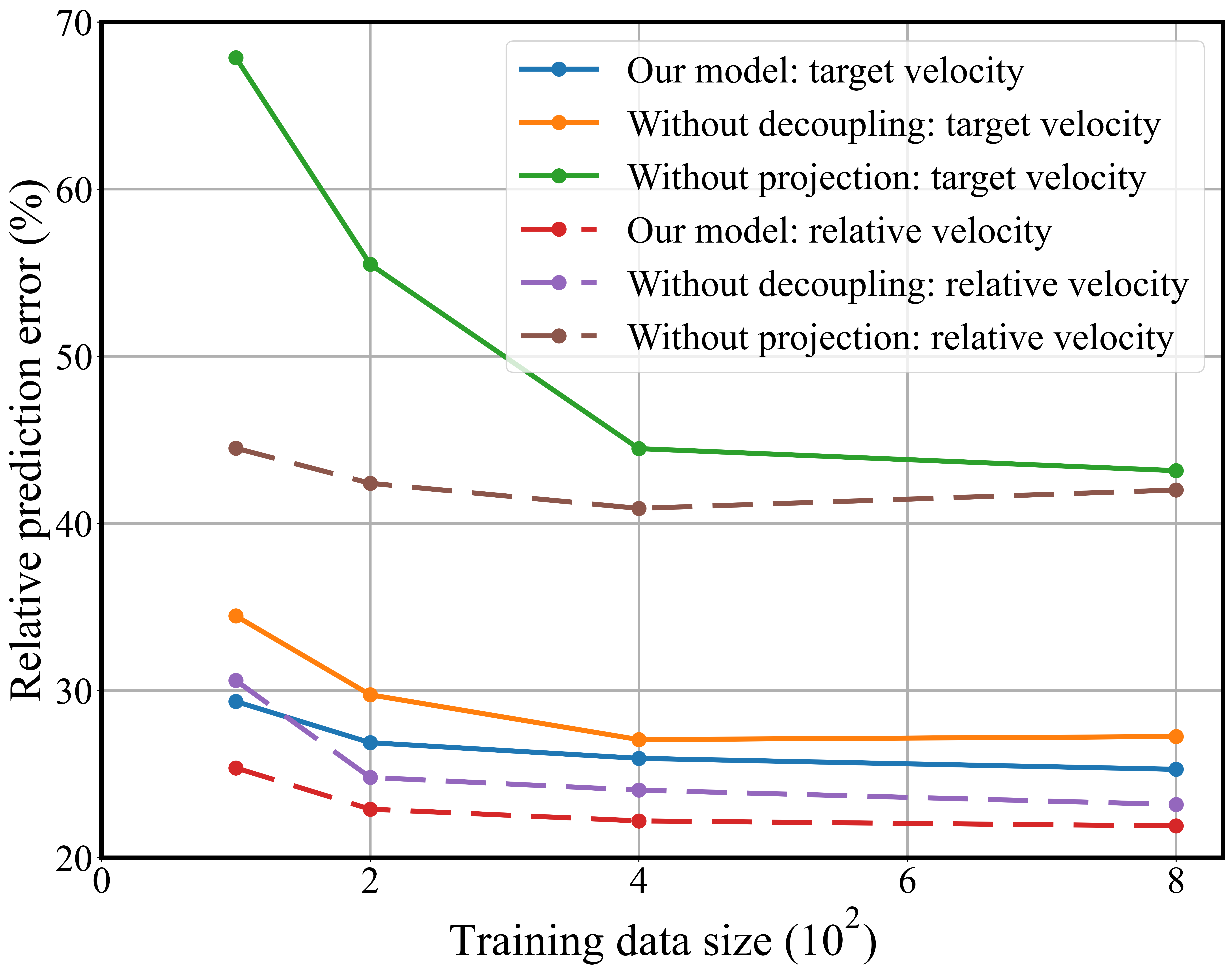}
    \vspace{-5pt}
    \caption{Validation of the data efficiency of local coordinate projection and decoupling method}
    \label{fig:model error}
    \vspace{-10pt}
\end{figure}

We first present a series of comparative studies illustrating the efficiency and real-time performance of the proposed model estimation method. Offline data is collected manually using Labview in the real environment. Subsequently, the RBF network approximation model is trained using the method outlined in Section IV-A.  The relative prediction errors of $\dot{\bm{x}}_u$ and $\dot{\bm{x}}_r$ are defined as
\begin{equation}
    e^{\text{test}}_k = \frac{\|\dot{\bm{x}}_k - \hat{\bm{g}}_k(\bm{x}_r)\bm{u}\|_2}{\|\dot{\bm{x}}_k\|_2} \times 100\%. \quad k\in\{u,r\}
\end{equation}
First, we evaluate the impact of different models and the amount of training data on the modeling performance. As shown in Fig. \ref{fig:model error}, the model corresponding to (\ref{eq: whole_linear}) exhibits the highest test error due to the absence of local coordinate system projection. Our model utilizing the decoupling of normal and tangential directions demonstrates superior data efficiency compared to the non-decoupled model. The relative modeling error under the training set of about 800 is about $20\%$, among the same order of magnitude as the Brownian motions (with an average displacement of 20 $\mu m$ per second according to the Stokes-Einstein equation), validating the efficiency of our modeling approach.
\begin{figure*}[ht]
    \centering
    \includegraphics[width=14cm]{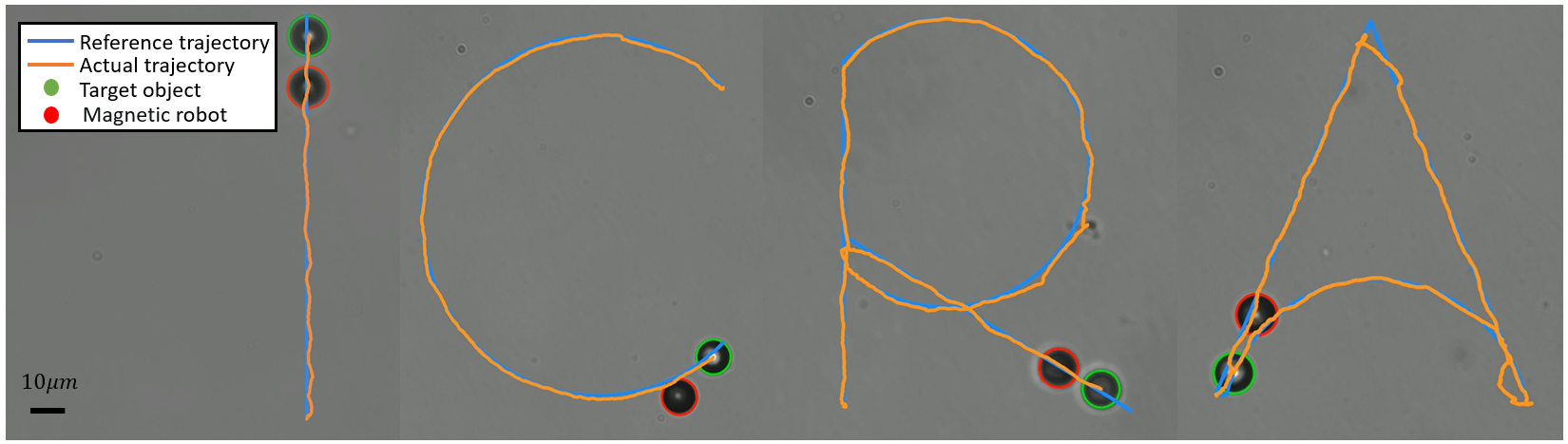}
    \vspace{-5pt}
    \caption{Trajectory tracking of ``ICRA". The green circle represents the target object, the red circle represents the magnetic-driven robot, the blue line represents the reference trajectory, and the orange line represents the actual trajectory.}
    \label{fig:icra}
    \vspace{-15pt}
\end{figure*}
\begin{table}[]
\caption{Activation Functions and Neuron Numbers}
\label{tab: exp success rate and node num}
\resizebox{\linewidth}{!}{
    \begin{threeparttable}
    
        \begin{tabular}{cccccc}
        \toprule[1pt]
        \multicolumn{2}{c}{{Neurons number}}   & {8}       & {16}      & {32}       & {64}       \\ 
        \midrule[0.8pt]
        \multirow{2}{*}{${e}_u^{\text{test}}$ $\downarrow$}      
                                      & Multi-quadric  & 0.192   & 0.198  & $\textbf{0.186}$   & 0.228   \\ \cline{2-2}
                                      & Gaussian  & 0.192   & $\textbf{0.186}$   & 0.188   & / \tnote{1} \\ \cline{1-2}
        \multirow{2}{*}{${e}_r^{\text{test}}$ $\downarrow$}      
                                      & Multi-quadric   & 0.222   & 0.226  & $\textbf{0.213}$   & 0.222  \\ \cline{2-2}
                                      & Gaussian   & 0.219   & 0.215  & 0.221   & 0.214   \\ 
       \bottomrule[1pt] 
       \end{tabular}
        \begin{tablenotes}
            \footnotesize
            \item[1] Gradient vanishing during the training process.
        \end{tablenotes}
    \end{threeparttable}
    
    }
    \label{tab: learning analyze}
    \vspace{-20pt}
\end{table}

We then compare the impact of two activation functions and the number of hidden layers of neurons on modeling accuracy. As summarized in Table \ref{tab: learning analyze}, the lowest prediction error is achieved with 32 neurons, ensuring the suitability of RBFN for real-time control. The multi-quadric activation function is ultimately chosen for its stability during training and smoothness of input.


\subsection{Validation of Trajectory Tracking Method}
To validate the dexterity of non-contact manipulation and the effectiveness of the proposed control method, we first designed trajectories involving the tracking of the four letters ``ICRA", which includes both straight-line segments and curves with different curvatures. The hyperparameters for the controller are set as $\alpha = 1.0$, $\lambda = 2.0$, $u_{\text{max}} = 2\pi(rad/s)$, $s_r^{\text{min}} = 2.25$, and $s_r^{\text{max}} = 4$, those for the online updater are $\tau_1 = \tau_2 = 0.1$. As shown in Fig. \ref{fig:icra}, the controller can successfully complete the tracking of all four trajectories while maintaining a non-contact state. The average tracking error for the letter ``C" is minimized, approximately 0.4 $\mu m$, with the maximum error occurring at the non-differentiable corner of the letter ``A", measuring 7.3 $\mu m$.

\begin{figure}[ht]
    \vspace{-5pt}
    \centering
    \includegraphics[width=6cm]{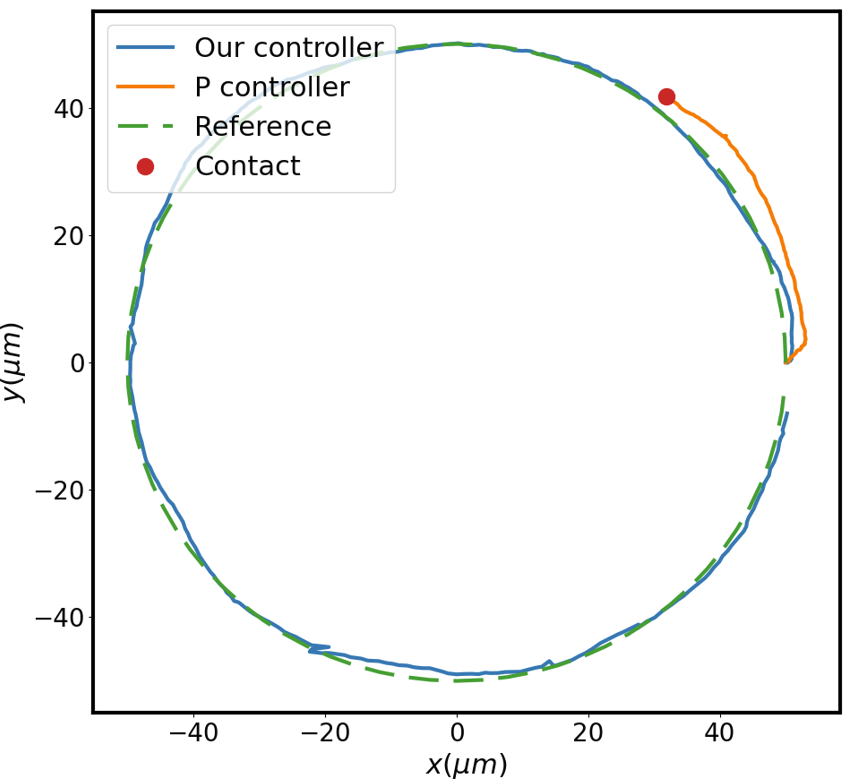}
    \vspace{-5pt}
    \caption{Comparison of the model-based optimal controller with proportional controller. The proportional controller leads to task failure since the robot comes into contact with the target object.}
    \vspace{-5pt}
    \label{fig:traj}
\end{figure}

We compare our proposed model-based optimal controller with the model-free Proportional (P-) controller.  
The actual trajectories are shown in Fig. \ref{fig:traj}. Our method completes the task. However, the error of the P-controller fails to converge, further, the robot and the object come into contact at $\left[31.82\mu m,  41.80\mu m\right]$, leading to adhesion and ultimately resulting in task failure.

\begin{table}[ht]
\caption{Control Parameters and Trajectory Curvatures}
\vspace{-2pt}
\resizebox{0.9\linewidth}{!}
{
    
    \begin{threeparttable}
    
        \begin{tabular}{ccccc}
        \toprule[1pt]
        \multicolumn{2}{c}{\multirow{2}{*}{Mean relative error ($100\%$)$\downarrow$}}   & \multicolumn{3}{c}{$\lambda$}   \\ 
         &  & {1}       & {2}      & {5}            \\ 
        \midrule[0.8pt]
        \multirow{3}{*}{Radius ($\mu m$)}      
                                      & 25  & 5.77   & 2.39  & $\textbf{1.83} $    \\ 
                                      & 50  & 0.626   & $\textbf{0.378}$   & 0.409 \\
                                      & 75 & 0.192 & $\textbf{0.186}$ &  0.188\\  
       \bottomrule[1pt] 
       \end{tabular}
    \end{threeparttable}
}
    \label{tab: radius}
\end{table}
\vspace{-5pt}
We further investigate the influence of control parameter $\lambda $ and curvature on tracking error for circular trajectories. Curvature is calculated by $\frac{1}{R}$. The average tracking error is recorded as TABLE \ref{tab: radius}. As the curvature increases, the relative tracking error increases, and the coefficient $\lambda$ should accordingly increase to achieve better tracking performance.
\begin{table}[t]
\caption{Comparison Between Different Planners}
    \begin{threeparttable}
    \resizebox{0.9\linewidth}{!}
    {\begin{tabular}{ccc}
        \toprule[1.2pt]
        {Method}  & Maximum curvature   $\downarrow$   & Maximum tracking $\downarrow$ \\
        & (rad/pixel) & error ($\mu m$)  \tnote{1}\\
        \midrule[0.8pt]
        CO-RRT     &  $\textbf{0.0084}$  &    $\textbf{2.654}$   \\ 
        Non-smooth  & 0.0532 & 5.320 \\  
        RRT  & 0.0827 &  14.367\\  
       \bottomrule[1.2pt] 
     \end{tabular}}
        \begin{tablenotes}
            \footnotesize
            \item[1] Due to the real-time variability of the experimental environment, tracking errors are tested under uniform non-obstacle conditions.
        \end{tablenotes}
    \end{threeparttable} 
    \label{tab: planner}
    \vspace{-20pt}
\end{table}

Finally, we validate the effectiveness of navigation in cluttered scenarios by comparing our planner with the planner without smoothing and the original RRT method. As shown in TABLE \ref{tab: planner}, our method generates a trajectory with the highest curvature, leading to the smallest tracking error. Additional navigation results are provided in the supplementary video.


\section{Conclusions}
This work presents a novel model, control, and navigation strategies for magnetic-driven non-contact micromanipulation. We introduce a decoupled affine non-linear model to delineate the dynamics of fluid-force effection subject to external magnetic field input. Subsequently, the neural networks are employed to efficiently learn the model from data and update weights online to enhance convergence speed. We propose an optimal controller based on the aforementioned model, which incorporates constraints to ensure non-contact manipulation. Finally, a curvature optimization planner is employed to guarantee navigation in clutter scenarios. The application of these techniques allows for precise control of target objects in a non-contact manner, facilitating trajectory tracking and reaching specific positions, without causing any physical damage. Future work will focus on applying this technology to single-cell manipulation.

{\small
\bibliographystyle{IEEEtran}
\bibliography{ref}
}


\end{document}